\documentclass{article}

\PassOptionsToPackage{numbers}{natbib}
\usepackage[final]{lmca2020}

\usepackage[utf8]{inputenc} % allow utf-8 input
\usepackage[T1]{fontenc}    % use 8-bit T1 fonts
\usepackage{hyperref}       % hyperlinks
\usepackage{url}            % simple URL typesetting
\usepackage{booktabs}       % professional-quality tables
\usepackage{amsfonts}       % blackboard math symbols
\usepackage{nicefrac}       % compact symbols for 1/2, etc.
\usepackage{microtype}      % microtypography

\usepackage{amsmath,amsthm,amssymb}
\usepackage{bm}
\usepackage{color}
\usepackage{subfigure}
\usepackage{dsfont}
\usepackage{blkarray, bigstrut}
\usepackage{graphicx}
\usepackage{makecell}
\usepackage{wrapfig}
\usepackage{enumitem}
\usepackage{algorithm}
\usepackage{algorithmic}
\usepackage{multicol}
\usepackage{multirow}
\usepackage{makecell}
\usepackage{comment}

\usepackage{thmtools}
\usepackage{thm-restate}

\usepackage{xcolor}
\definecolor{midgreen}{rgb}{0.1,0.5,0.1}
\definecolor{darkgray}{gray}{0.25}
\definecolor{lightblue}{rgb}{0.25,0.25,1}
\usepackage[capitalise]{cleverref}
\creflabelformat{equation}{#2#1#3}

\theoremstyle{definition}

\newcommand{\norm}[1]{\ensuremath{\left\| #1 \right\|}}

\def\argmax{\mathop{\rm argmax}}
\def\argmin{\mathop{\rm argmin}}

\def\GS{[\![M]\!]}

\def\0{{\bm 0}}

\def\c{{\bm c}}

\def\e{{\bm e}}

\def\x{{\bm x}}
\def\y{{\bm y}}
\def\z{{\bm z}}

\def\B{{\bm B}}
\def\C{{\bm C}}

\def\I{{\bm I}}

\def\K{{\bm K}}
\def\L{{\bm L}}

\def\S{{\bm S}}

\def\V{{\bm V}}

\def\Lambda{\boldsymbol{\lambda}}

\def\Jcal{\mathcal{J}}

\def\Ycal{\mathcal{Y}}

\def\Rbb{\mathbb{R}}

\title{Wasserstein Learning of Determinantal Point Processes}

\author{%
  Lucas Anquetil \thanks{Currently at INSA Rouen.} \\
  Criteo AI Lab \\
  \texttt{lucas.anquetil@insa-rouen.fr } \\
  \And
  Mike Gartrell \\
  Criteo AI Lab \\
  \texttt{m.gartrell@criteo.com} \\
  \And
  Alain Rakotomamonjy \\
  Criteo AI Lab \\ and University of Rouen \\
  \texttt{a.rakotomamonjy@criteo.com} \\
  \And
  Ugo Tanielian \\
  Criteo AI Lab \\ and Sorbonne University \\
  \texttt{u.tanielian@criteo.com} \\
  \And
  Clément Calauzènes \\
  Criteo AI Lab \\
  \texttt{c.calauzenes@criteo.com} \\
}

\definecolor{Blue}{rgb}{0.0,0.0,1.0}

\begin{document}

\maketitle

\vspace{-0.4cm}
\begin{abstract}
\vspace{-0.2cm}
Determinantal point processes (DPPs) have received significant attention as an elegant probabilistic model for discrete subset selection.  Most prior work on DPP learning focuses on maximum likelihood estimation (MLE).  While efficient and scalable, MLE approaches do not leverage any subset similarity information and may fail to recover the true generative distribution of discrete data. In this work, by deriving a differentiable relaxation of a DPP sampling algorithm, we present a novel approach for learning DPPs that minimizes the Wasserstein distance between the model and data composed of observed subsets. Through an evaluation on a real-world dataset, we show that our Wasserstein learning approach provides significantly improved predictive performance on a generative task compared to DPPs trained using MLE.
\end{abstract}

\vspace{-0.5cm}
\section{Introduction}
\vspace{-0.2cm}
%The recommendation task of basket generation is a key part of many online retail applications. It involves learning from existing basket of products gathered from 

%GANs \citep{GANs} are an effective way to learn complex and high-dimensional distributions, leading to state-of-the-art models for image synthesis in both unconditional \citep{karras2018style} and conditional settings \citep{brock2018large}. However, it is well-known that in the discrete setting, GANs are shown to be highly complicated to train. A major issue of GANs when applied to discrete settings involves the complexity of back-propagating the gradients for the generative model. %\citet{kusner2016gans} were the first to define a sampling scheme with the use of a Gumbel softmax distribution. 
%To bypass this differentiation problem,~\citet{SeqGANs} proposed a cost function inspired by reinforcement learning. In this setting, the discriminator is used as a reward function, and the generator is trained via policy gradient \citep{sutton2000policy}. 

%In the current paper, we propose an elegant way to apply GANs ...

Generative models have enjoyed a great deal of success in the recent years due to their
ability to capture insights from data distributions.  
Those models have generally been applied to continuous data by training them using maximum likelihood estimation (MLE) or, more recently, using adversarial learning with the well-known Generative Adversarial Networks framework~\cite{GANs, karras2018style}.

When dealing with discrete data, generative models trained with MLE suffer from a bias due to the asymmetrical definition of MLE. Equivalent to minimizing a Kullback Leibler divergence, the MLE cost function pays extremely low cost for generating low-quality samples. Consequently, a generative model trained by MLE tends to cover the full data distribution at the expense of covering unnecessary regions~\cite{chao2015large, mariet2019learning}. 
On the other hand, when considering adversarial learning of discrete generative models, one usually exploits the gradient of the discriminator's loss when optimizing the generator. However, since the gradient computation requires backpropagation through the generator's output, i.e. the data, adversarial approaches are difficult to apply when generating discrete data. Depending on the generative model and the structure of the data, there are some ways to overcome this issue. For instance, \cite{kusner2016gans} was the first to define a sampling scheme with the use of a Gumbel softmax distribution, and several generalizations of this softmax trick have been recently proposed in the literature~\cite{paulus2020gradient,joo2020generalized}. 

In this work, we address the problem of training a determinantal point process (DPP), a probabilistic model for subsets drawn from a large collection of items. A DPP parameterizes a probability distribution over the combinatorial space of subsets of elements drawn from $\Jcal$, which is a discrete space composed of $M$ distinct items. DPPs are appealing models for this setting, since they are known to also capture interactions between elements within subsets. More importantly, they offer efficient polynomial-time algorithms for most probabilistic inference operations over the space of $2^M$ possible subsets, such as normalization, learning, and sampling~\cite{gillenwater2014approximate, kulesza2012determinantal}. In order to move away from the standard MLE learning framework~\cite{gartrell17, mariet2015}, which may suffer from the flaw described above, we define a new learning scheme for DPPs based on the minimization of the Wasserstein distance between the samples generated by the DPP and the training data. Compared to MLE, one of the main benefits of this Wasserstein-based approach is that it allows us to define a transportation cost function (\emph{e.g} a Jaccard distance) that induces a bias on the assumed structure of the space of subsets. Minimizing this cost function allows the learning to take into account differences between pairs of subsets and to reduce the distance between subsets based on their similarities. We argue that this Wasserstein-based scheme leverages more information from the data and results in a better approximation of the target distribution. 

The contributions of this work are the following: \textbf{1.)} We present a new framework when learning DPPs that minimizes the Wasserstein distance between the DPP and data composed of observed subsets. This framework can be applied to any generative probabilistic model for discrete sets. \textbf{2.)} Leveraging recent work on a DPP sampling algorithm with computational complexity that is sublinear in the size of the ground set~\cite{derezinski2019exact}, and stochastic softmax tricks for gradient estimation of discrete distributions~\cite{joo2020generalized}, we present a differentiable DPP sampling algorithm that can scale to large ground sets.
\textbf{3.)} We evaluate our Wasserstein learning approach on a real-world dataset, and show substantial improvements in predictive performance compared to DPPs trained using MLE. This experimental evaluation is one of the first to focus on a generative modeling task for DPPs.

\vspace{-0.2cm}
\section{Background and related work}
\label{sec:background}
\vspace{-0.2cm}
\paragraph{\textbf{Determinantal Point Processes}}
Consider a finite set $\Jcal=\{1,2,\ldots,M\}$ of cardinality $M$, which we will also denote by $\GS$.  A DPP defines a probability distribution over all $2^M$ subsets.  It is parameterized by a matrix $\L \in \Rbb^{M \times M}$, called the \textit{kernel}, such that the probability of each subset $J \subseteq \GS$ is proportional to the determinant of its corresponding principal submatrix: $\Pr(Y) \propto \det(\L_J)$, where $\L_J=[\L_{ij}]_{i,j\in J}$ is the submatrix of $\L$ indexed by $J$.  The normalization constant for this distribution can be expressed as a single $M \times M$ determinant: $\sum_{J \subseteq \GS}\det(\L_J) = \det(\L + \I)$ \citep[Theorem 2.1]{kulesza2012determinantal}.  Therefore, $\Pr(J) = \det(\L_J) / \det(\L + \I)$.

In order to ensure that the DPP defines a probability distribution, all principal minors of $\L$ must be non-negative: $\det(\L_J) \geq 0$.  Matrices that satisfy this property are called $P_0$-matrices~\cite[Definition 1]{fang1989laa}.  Several decompositions of $\L$ that partially cover the $P_0$ space are known. One common decomposition that covers the space of symmetric $P_0$-matrices exploits the fact that $\L \in P_0$ if $\L$ is positive semidefinite (PSD)~\cite{prussing1986jgcd}.  Any symmetric PSD matrix can be written as the Gramian matrix of some set of vectors: $\L := \V \V^\top$, where $\V \in \mathbb{R}^{M \times K}$.  We restrict our work in this paper to such symmetric DPPs with this decomposition, since efficient sampling algorithms, such as~\cite{derezinski2019exact}, are only available for symmetric DPPs.  There are decompositions of $\L$ that partially cover the nonsymmetric $P_0$~\cite{gartrell2019nonsym}; we leave an investigation of Wasserstein learning of nonsymmetric DPPs for future work.

In this work we use the DPP-VFX sampling algorithm~\cite{derezinski2019exact}, which has computational complexity sublinear in $M$, and is therefore one of the most efficient exact sampling methods for DPPs. DPP-VFX relies on a connection between ridge leverage scores~\cite{alaoui2015fast} and DPPs to implement a distortion-free intermediate sampling method that enables this sublinear time complexity.  Since DPP-VFX requires a base DPP sampling algorithm, we propose to use the Cholesky-based DPP sampling approach~\cite{launay2020exact, poulson2020high}.

\vspace{-0.2cm}
\paragraph{\textbf{Estimating gradients in discrete settings}}
The Wasserstein learning approach requires computing gradients over discrete subset samples drawn from a DPP. Two families of approaches for discrete gradient estimation are score function estimators, such as REINFORCE~\cite{williams1992simple}, and continuous relaxations of discrete distributions, most of which are based on the Gumbel-Max trick~\cite{maddison2014sampling}.  REINFORCE has the drawback of high variance, making it impractical in many cases.  While techniques for variance reduction exist~\cite{mnih2014neural}, they often involve highly engineered control variates.  Relaxed gradient estimators incorporate bias in order to reduce variance, and are often easier to implement~\cite{paulus2020gradient}.  We choose the relaxation approach, and leverage recent work on stochastic softmax tricks~\cite{kusner2016gans, paulus2020gradient}, which is a unified framework for structured relaxations of discrete combinatorial distributions.  In particular, we use stochastic softmax tricks to develop a differentiable version of the DPP-VFX sampling algorithm, with a differentiable version of the Choleksy-based approach as the base DPP sampling algorithm.  As far as we are aware, this is the first instance of a differentiable DPP sampling algorithm.
% which may be of independent interest.

\vspace{-0.2cm}
\section{Learning DPPs via Wasserstein minimization}
\vspace{-0.2cm}
The classical approach for learning a DPP kernel given a collection of subsets is to maximize the likelihood of data samples drawn from the same distribution as the one used for obtaining training examples~\cite{gartrell17, gartrell2019nonsym}. One advantage of optimizing the (log) likelihood is that the likelihood of samples has a closed form expression with respect to the model parameters. Since that expression is continuously differentiable, a gradient ascent algorithm is a natural solution for solving the problem. Instead of likelihood maximization, we propose a DPP learning approach that minimizes the Wasserstein distance between the training data and samples generated by the model. This optimization scheme seeks to improve the approximation of the generative distribution of the data. 

The Wasserstein distance is a distance between probability distributions defined on a given metric space. We let $\mathcal{X}_n =\{x_1, \cdots, x_n\}$ denote the training dataset of size $n$ with empirical distribution $\mu = \sum_{i=1}^n a_i \delta_{\x_i}$, where $\delta$ refers to the Dirac distribution, and $\Ycal_n=\{y_1, \cdots, y_n\}$ be the collection of $n$ sets sampled from the DPP model with distribution $\nu = \sum_{i=1}^m b_i \delta_{\y_i}$, where the $a_i$ and $b_i$ follow a uniform distribution. Given a transportation cost $d$ defined on $2^M \times 2^M $, the Wasserstein distance between $\mu$ and $\nu$ seeks an optimal coupling $P$ defined on $[1,n]^2$ that minimizes the cost of transporting mass from $\mu$ to $\nu$ \cite{peyre2019computational}. When dealing with discrete sets of items, we argue that the use of the Jaccard distance~\cite{kosub2019note} as as transportation cost function is a good choice. The Jaccard distance between two sets takes into account both the difference in length and in the items chosen: $d_J(X, Y) = (|X \cup Y| - |X \cap Y|)/|X \cup Y|$,
where $X, Y \in 2^M$. Since the cost function needs to be differentiable, we use a differentiable proxy for Jaccard distance. For $x \in \mathcal{X}_n$ and $y \in \Ycal_n$, the differentiable Jaccard distance $d_S$ is defined as follows:
\begin{equation}\label{eq:proxy_jaccard}
    d_S(\x, \y) = 1 - \frac{\x^\top \y}{M - (1 - \x)^\top (1 - \y)} \;,
\end{equation}
where $\x, \in \{0, 1\}^M$ is a binary indicator vector and $\y \in [0, 1]^M$ is a continuous relaxation of a binary vector, with $y_k, k \in [1,M]$ being the inclusion probability of item $k$ in the sample. By combining the definition of the Wasserstein distance with the chosen cost function in \eqref{eq:proxy_jaccard}, we define the following Wasserstein optimization problem for DPPs:
\begin{equation}
    \argmin_{\substack{\V \in \mathbb{R}^{M \times K}}} 
    %\min_{P \in \Pi(\Bcal,\Ccal)}
    \sum_{\substack{i,j=1}}^n P_{i,j}^\star d_S(\x_i, \y_j) + \alpha \norm{\V}_F^2 \quad  \text{with} \quad  P^\star= \argmin_{P \in \Pi(\mu,\nu)}  \sum_{\substack{i,j=1}}^n P_{i,j} d_S(\x_i, \y_j) \;,
    \label{eq:wasserstein-opt}
\end{equation}
where $\{\x_1, \cdots, \x_n\}$ is the training data, $\{\y_1, \cdots, \y_n\}$ is a collection of $n$ subsets drawn from the DPP, and $\alpha \ge 0$ is a tunable hyperparameter for regularization.  Recall that we use the decomposition $\L = \V \V^\top$ for the DPP kernel.

\begin{wrapfigure}{L}{0.5\textwidth}
\vspace{-0.8cm}
\begin{minipage}{0.5\textwidth}
\begin{algorithm}[H]
\caption{Wasserstein learning} 
\label{alg:wd-learning}
\begin{algorithmic}
\STATE {\bf Input:} training data, $\V \in \mathbb{R}^{M \times K}$, maxIter
\FOR {maxIter steps}
    \STATE \textbf{Sample} subsets from training data.
    \STATE \textbf{Sample} subsets from DPP.
    \STATE \textbf{Compute} $P^\star$ in \cref{eq:wasserstein-opt} with \cite{flamary2017pot}.
    \STATE  \textbf{Update} $\V$ using \cref{eq:wasserstein-opt}
    %\STATE \textbf{Update} $\V$ with \cref{eq:wasserstein-opt}. 
    %$\V_{i+1} \leftarrow$ \textbf{OptimizeWD}$(\V_i, \Bcal, \Ccal)$ (\cref{eq:wasserstein-opt})
\ENDFOR
\end{algorithmic}
\end{algorithm}
\end{minipage}
\end{wrapfigure}
For the sake of completeness, the algorithm used to solve this optimization scheme is described in \cref{alg:wd-learning}. Solving the optimization problem defined in \cref{eq:wasserstein-opt} with backpropagation requires computing the gradient on minibatches with respect to the parameters $\V$, and thus a differentiable sampling algorithm is needed. We use an estimation of the Wasserstein distance on a minibatch \cite{fatras2019learning} by computing a Earth Mover distance, and consider a new differentiable formulation of the DPP-VFX sampling algorithm~\cite{derezinski2019exact}, shown in \cref{alg:vfx-sampling} in \cref{sec:diff-dpp-vfx}. We apply the Gumbel softmax trick to the base Poisson, and multinomial, and Bernoulli sampling steps in DPP-VFX. Combined with a differentiable  DPP Cholesky-based sampler (\cref{alg:differentiable-cholesky}), this sampler generates continuous relaxations of binary indicator vectors for subsets; see \cref{sec:diff-dpp-vfx} for details. 
%If we replace the DPP with another generative model for discrete subsets, and use a differentiable sampling algorithm for this model, our Wasserstein learning approach can be easily applied to other families of models.  We leave an investigation of our approach for other models for future work.
\vspace{-0.2cm}
\section{Experiments}
\vspace{-0.2cm}
We perform experiments on the Amazon Baby Registries dataset.  This dataset consists of registries or "baskets" of baby products, and has been used in prior work on DPP learning~\cite{gartrell2016bayesian, gartrell2019nonsym, gillenwater14, mariet2015}.  The registries contain items from 15 different categories, such as ``apparel'', with a catalog of up to 100 items per category.  We evaluate on the most popular apparel category, which contains 14{,}970 registries, as well as the popular diaper and feeding categories.  
% We leave experiments on larger datasets for future work.

\vspace{-0.25cm}
\subsection{Setup and evaluation metrics}
\label{subsec:experimental-setup}
\vspace{-0.25cm}
A small set consisting of 300 randomly-selected baskets is kept for validation, and a further random selection of 2000 baskets is used for testing. We implement our models using PyTorch~\cite{paszke2019pytorch}; Adam~\cite{kingma2015adam} is used for optimization, in conjunction with the solver from the POT package~\cite{flamary2017pot}. 

We use the low-rank symmetric DPP (SDPP)~\cite{gartrell17} and the low-rank nonsymmetric DPP (NDPP)~\cite{gartrell2019nonsym}, both trained using MLE, as baseline models for all experiments. We evaluate these baselines and our Wasserstein DPP model (WDPP) model on a subset generation task, where we estimate the Wasserstein distance (WD) between subsets sampled from the model and subsets in the test set by computing the Earth Mover's distance between these two subset collections using POT~\cite{flamary2017pot}.
%\begin{itemize}
%    \item Subset Generation task: we measure the Wasserstein distance between the DPP distribution and the target distribution. This distance is estimated by computing the Earth Mover's distance between collections of sets. This is done via the use of the POT package~\cite{flamary2017pot}.
%    \item Subset Discrimination task: we test the discriminative ability of a model to distinguish between sets in the test data and random ones. The score for each subset is the log-likelihood that assigned by the model and report the AUC in \cref{tab:predictive-qual}. We also perform a variant of this task, where we randomly remove an item from a set observed in the test data. 
    %For each subset in the test set, we generate a subset of the same length by drawing items uniformly at random (and we ensure that the same item is not drawn more than once for a subset) \ar{This is not very clear for me. I thought the idea was to remove/replace only one item at random... }.  %We also perform a variant of this task, where we randomly select a subset composed of half of the elements within each subset in the test set, and then generate a random subset of the same length; results for this task are shown under ``AUC (subsets)'' in \cref{tab:predictive-qual}.
%\end{itemize}

\begin{table}[t]
  \caption{Wasserstein distance (WD), and test log-likelihood (test ll) for all datasets, for the symmetric DPP (SDPP), nonsymmetric DPP (NDPP), and the Wasserstein DPP (WDPP). WD results show 95\% confidence estimates obtained via bootstrapping. Bold values indicate the best performance. 
  }
  \vspace{-0.2cm}
  \centering
  \scalebox{0.77}{
    \begin{tabular}{cccccccccc}
      \multicolumn{4}{c}{\thead{Amazon: Apparel ($M = 100$)}} &
      \multicolumn{3}{c}{\thead{Amazon: Diaper ($M = 100$)}} & 
      \multicolumn{3}{c}{\thead{Amazon: Feeding ($M = 100$)}} \\
      Metric & SDPP & NDPP & WDPP & SDPP & NDPP & WDPP & SDPP & NDPP & WDPP \\
      \hline
      WD & 0.76 ${\scriptstyle \pm 0.01}$ & 0.76  ${\scriptstyle \pm 0.01}$ & \bf 0.58  ${\scriptstyle \pm 0.01}$ & 0.72 ${\scriptstyle \pm 0.01}$ & 0.73 ${\scriptstyle \pm 0.01}$ & \bf 0.63  ${\scriptstyle \pm 0.01}$ & 0.69 ${\scriptstyle \pm 0.01}$ & 0.69 ${\scriptstyle \pm 0.01}$ & \bf 0.65 ${\scriptstyle \pm 0.01}$ \\
      Test ll & -10.09 &  -9.60 & -17.78 & -10.54 &  -9.98 & -14.27 & -12.13 & -11.67 & -17.65 \\ 
    \hline
    \end{tabular}
    }
  \label{tab:predictive-qual}
\end{table}

\begin{figure}
    \vspace{-0.1cm}
    \centering
    {\includegraphics[scale=0.37]{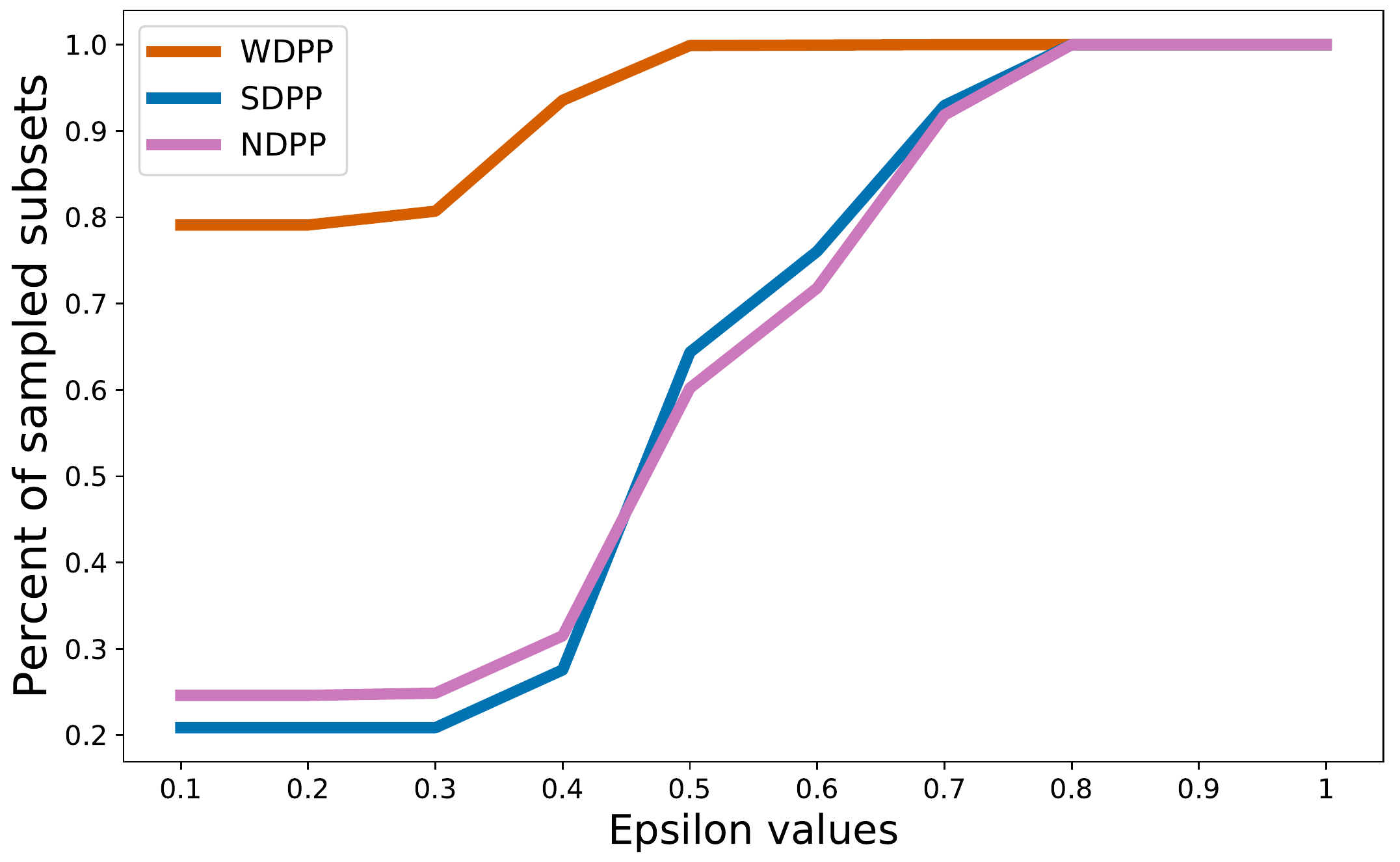}}
    \vspace{-0.3cm}
    \caption{Precision plot for the generated subsets from each model that have a Jaccard distance of at most $\epsilon$ with at least one subset in the test set, for $\epsilon \in (0, 1]$, for the Amazon apparel dataset.}
    \label{fig:precision-plot}
    \vspace{-0.5cm}
\end{figure}

\vspace{-0.25cm}
\subsection{Results}
\vspace{-0.25cm}
\label{subsec:experimental-results}
Consistent with prior work, we see that the MLE NDPP outperforms the MLE SDPP on the test log-likelihood metric.  However, we also observe that MLE is not directly connected to the generative task, and higher performance on test log-likelihood does not result in higher performance on the WD metric.  As expected, since the Wasserstein learning approach directly optimizes a proxy for the generative task, the WDPP model significantly outperforms the baseline models in terms of WD. To provide some evidence of the connection between the WD metric and the quality of generated subsets, \cref{fig:precision-plot} shows the percentage of generated subsets from each model that have a Jaccard distance of at most $\epsilon$ with at least one subset in the test set, for $\epsilon \in (0, 1]$. For any given $\epsilon$, we see that WDPP outperforms MLE models. This highlights that the WDPP, by being able to take the Jaccard distance into account, exploits the underlying structure of the combinatorial space $2^M$, while the MLE-trained models do not and thus treat all subsets as completely different.
%We see the WDPP significantly outperforms the SDPP and NDPP baselines, and therefore the WDPP generates subsets that are more realistic than the MLE baselines. % Regarding the partial subset discrimination task, we see that the WDPP model outperforms the DPP models trained using MLE.  This results provides some empirical evidence that the Wasserstein learning approach induces additional structure over the combinatorial set space, which is not available with MLE.  This additional structure appears to allow the WDPP model to more effectively learn relationships between items and allocate higher probability mass to partial subsets constructed from full subsets observed in the test set.
We present additional experimental results in~\cref{sec:additional-experiments}.  These results provide further evidence that, compared to MLE DPPs, the WDPP model recovers significantly more structure, and is able to generate subsets that are substantially closer to observed data.

\vspace{-0.25cm}
\section{Conclusion}
\vspace{-0.25cm}

We have presented a new Wasserstein learning approach for DPPs. Unlike conventional MLE learning approaches for DPPs, this learning approach optimizes a proxy for discrete subset generation. Empirical results indicate that the proposed approach leads to substantially improved generative performance compared to MLE. This approach is fully general, and can be readily applied to other families of models for discrete subsets. We leave such an investigation for future work.

\bibliographystyle{plainnat}
\setcitestyle{numbers}
\bibliography{bibliography}

\clearpage
\appendix

\section{Differentiable DPP-VFX Sampling Algorithm}
\label{sec:diff-dpp-vfx}

As indicated in \cref{sec:background}, we have leveraged stochastic softmax tricks (SST)~\cite{paulus2020gradient} to develop a differentiable version of the DPP-VFX sampling algorithm~\cite{derezinski2019exact}.  Compared to other DPP sampling algorithms, DFF-VFX can be substantially faster, since it has time complexity sublinear in $M$.  DPP-VFX uses a connection between ridge leverage scores~\cite{alaoui2015fast} and DPPs to implement a distortion-free intermediate sampling method that enables this sublinear time complexity.  The first step of the sampling algorithm downsamples the items in $\GS$ i.i.d. with probability proportional to the ridge leverage score of each item, and then runs a conventional DPP sampling algorithm on this thinned or downsampled set of items, whose cardinality is much smaller than $M$.  We use a differentiable version of the Cholesky-based DPP sampling algorithm~\cite{launay2020exact, poulson2020high} on this thinned set of items.  When downsampling the items, the first step is to select the number of items that will be kept using a Poisson sampling step, followed by a multinomial sampling step that selects the items that will be included in the downsampled set.  Finally, a Bernoulli sampling step is used to perform rejection sampling, in order to ensure that the final exact DPP sample will be contained within the downsampled set.  

Our differentiable DPP-VFX sampling algorithm is presented in \cref{alg:vfx-sampling}, where $\beta$ is a kernel rescaling parameter that ensures that the Poisson parameter $(s*\e^{s/q})$ is equal to the catalog size $M$.  Differentiable versions of the Poisson, multinomial, Bernoulli, and Cholesky-based samplers invoked by \cref{alg:vfx-sampling} are shown in \cref{alg:sst-poisson}, \cref{alg:sst-multinomial}, \cref{alg:differentiable-bernoulli}, and \cref{alg:differentiable-cholesky}, respectively.

% The main issue with the DPP Cholesky sampling algorithm is its time complexity, linear in the catalog size. Indeed using this algorithm, to sampled one set from the DPP, we need to go over all of the items and pick the ones that pass the sampling criterion. However there exists a sampling algorithm that is sublinear of the catalog size in its time complexity, the VFX sampling algorithm, that we implemented in a differentiable way. The main enhancement of the VFX to the Cholesky linear sampling, is that we don't need to go over all of the items when sampling one set. Actually the first step of the algorithm is to downsample the set of items according to their probability and then run the Cholesky sampling algorithm on the thinned set of items. When downsampling the items, the first step is to select the number of items that will be kept (the Poisson sampling), then we select the items that will define the downsampled set of items (the Multinomial sampling), finally it operates a Bernoulli sampling to confirm that the downsampled set of items define a good approximation of the original set distribution.

% In order to make everything differentiable we implemented the discrete samplings of the VFX algorithm (Poisson, Multinomial and Bernoulli) as Stochastic Softmax Trick differentiable functions following a concrete application of \cite{paulus2020gradient}.
\begin{algorithm}[h!]
\caption{Differentiable DPP-VFX sublinear sampling S $\sim$ DPP($\L$)} 
\label{alg:vfx-sampling}
\begin{algorithmic}
\STATE {\textbf{Input:}} $\L \in \mathbb{R}^{M\times M}$, 
$\beta > 0$
\STATE {\textbf{Initialization:}} $\L \leftarrow \beta * \L ,$ 
$\K \leftarrow \I - (\L + \I)^{-1}$
\STATE $l_i \leftarrow K_{ii} \approx Pr(i \in S) ,$ 
$s \leftarrow \sum_{i} l_{i}$ 
\STATE \bf{if} {$s > 1$} \bf{then} $q \leftarrow s^2$ \bf{else} $q \leftarrow s$
\STATE $\widetilde{\L} \leftarrow \frac{s}{q}[\frac{1}{\sqrt{l_{i}l{j}}}L_{i,j}]_{i,j} $

\STATE \textbf{Downsampling:} $Acc \leftarrow$ \textnormal{False}
\WHILE{$ \NOT Acc$}
    \STATE $t \sim$ \textnormal{SST-Poisson}($s*\e^{s/q})$ \textnormal{(\cref{alg:sst-poisson})}
    \STATE $\sigma_1,...,\sigma_t \stackrel{i.i.d}{\sim} $ \textnormal{SST-Multinomial}$\left( \frac{l_1}{s},...,\frac{l_n}{s} \right)$ \textnormal{(\cref{alg:sst-multinomial})}
    \STATE $Acc \sim$ \textnormal{SST-Bernoulli} $\left( \frac{\e^{s}\det(\I + \widetilde{\L}_{\sigma})}{\e^{ts/q}\det(\I + \widetilde{\L})} \right)$ \textnormal{(\cref{alg:differentiable-bernoulli})}

    \ENDWHILE
\STATE \COMMENT{\textnormal{Sample from thinned item catalog:}} $\widetilde{S} \sim \textnormal{DPP}(\widetilde{\L}_{\sigma})$ \textnormal{(\cref{alg:differentiable-cholesky})}
\STATE {\bf return} $\S = \{\sigma_i : i \in \widetilde{S}\}$
\end{algorithmic}
\end{algorithm}

\begin{algorithm}[h!]
\caption{SST-Poisson sampling} 
\label{alg:sst-poisson}
\begin{algorithmic}
\STATE {\bf Input:} $\lambda$, temperature $\tau$
\STATE {\bf STEP 1:} \COMMENT{Truncate the total support and compute the probabilities of the integers from $1$ to $2*\lambda$}
\STATE massLogProb $\leftarrow$ $\log(\text{Poisson}_{\lambda}(i))$ for i in $[0,...,2*\lambda]$
\STATE {\bf STEP 2:} \COMMENT{Differentiable sampling using the Gumbel Softmax trick over the massLogProb log mass probability distribution}
\STATE oneHotSample $\sim$ $\text{GumbelSoftmax}_{\tau}(\text{massLogProb})$
\STATE {\bf STEP 3:} \COMMENT{Rearrange the one-hot-vector sample into the desired output format using matrix operations}
\STATE sstPoissonSample $\leftarrow \I \cdot$ oneHotSample
\STATE {\bf return} sstPoissonSample
\end{algorithmic}
\end{algorithm}

\begin{algorithm}[h!]
\caption{SST-Multinomial sampling} 
\label{alg:sst-multinomial}
\begin{algorithmic}
\STATE {\bf Input:} masslogprob, nbsample, upperbound, temperature $\tau$
\STATE {\bf STEP 1:} \COMMENT{Sample upperbound differentiable multinomial samples}
\STATE allSamples $\leftarrow$ [$\text{GumbelSoftmax}_{\tau}(\text{masslogprob})$] \textbf{for} i \textbf{in} [0,...,upperbound]
\STATE {\bf STEP 2:} \COMMENT{Select nbsample unique samples from allSamples}
\STATE uniqueMultinomialSamples $\leftarrow$ unique(allSamples)
\STATE {\bf return} uniqueMultinomialSamples
\end{algorithmic}
\end{algorithm}

\begin{algorithm}[h!]
\caption{SST-Bernoulli sampling} 
\label{alg:differentiable-bernoulli}
\begin{algorithmic}
\STATE {\bf Input:} value, temperature $\tau$
\STATE {\bf STEP 1:} Sample a uniform value and build a massLogProb out of the two values
\STATE randValue $\leftarrow$ uniform(0, 1)
\STATE massLogProb $\leftarrow$ $[\log(\text{value}), \log(\text{randValue})]$, \COMMENT{massLogProb $\in \mathbb{R}^2$}
\STATE {\bf STEP 2:} \COMMENT{Differentiable sampling using the Gumbel Softmax trick over the massLogProb log mass probability distribution}
\STATE oneHotSample $\leftarrow$ $\text{GumbelSoftmax}_{\tau}(\text{massLogProb})$
\STATE {\bf return} $\text{oneHotSample[0]}$
\end{algorithmic}
\end{algorithm}

% The VFX is built upon the linear Cholesky sampling algorithm, that was implemented in a differentiable way too :
\begin{algorithm}[h!]
\caption{Differentiable DPP Cholesky linear sampling S $\sim$ DPP($\L$)} 
\label{alg:differentiable-cholesky}
\begin{algorithmic}
\STATE {\bf Input:} $\L \in \mathbb{R}^{M\times M}$, temperature $\tau$
\STATE $\K \leftarrow \I - (\L + \I)^{-1}$
\STATE $ S \leftarrow []$
\FOR{\textbf{each} item $i$ \textbf{in} catalog} {
\STATE $\text{itemValue} \sim $Differentiable-Bernoulli($K_{\text{i,i}}$) (\cref{alg:differentiable-bernoulli})
\STATE \COMMENT{Add 0 or soft-value to S:} 
\STATE S $\leftarrow$ S + $\text{binary}(\text{itemValue}) * \text{sigmoid}(\text{itemValue}/\tau)$
\STATE \COMMENT{Update the kernel according to the item sample:}
\STATE $K_{\text{i,i}} \leftarrow K_{\text{i,i}} - (1 - \text{binary}(\text{itemValue}) )$

\STATE $K_{\text{[i+1:M],i}} \leftarrow K_{\text{[i+1:M],i}} / K_{\text{i,i}}$
\STATE $K_{\text{[i+1:M],[i+1:M]}} \leftarrow K_{\text{[i+1:M],[i+1:M]}} - K_{\text{[i+1:M],i}} \otimes K_{\text{i,[i+1:M]}}$
}
\ENDFOR
\STATE {\bf return} $S$
\end{algorithmic}
\end{algorithm}

\subsection{Gumbel-Softmax trick}
Our differentiable DPP sampling approach relies on the Gumbel-Softmax reparameterization trick~\cite{jang2016categorical}, which is an efficient gradient estimator that replaces the non-differentiable sample from a discrete distribution with a differentiable sample from a Gumbel-Softmax distribution.  The Gumbel-Max trick provides a simple and efficient way to draw samples $\z$ from a discrete distribution with class probabilities $\pi_{i}$ : 
\begin{equation}
    \z = \text{oneHot}\left(\argmax_{i}[g_{i} + \log(\pi_{i})]\right)
\end{equation}
where $g_{1}...g_{k}$ are i.i.d samples drawn from Gumbel(0, 1). The softmax is used as a continuous, differentiable approximation to $\argmax$, and generates $k$-dimensional sample vectors $\y \in \Delta^{k-1}$, where each component $y_i$ is: 
\begin{equation}
    y_{i} = \frac{\exp((\log(\pi_{i}) + g_{i})/\tau)}{\sum_{j=1}^{k} \exp((\log(\pi_{j}) + g_{j})/\tau} \text{ for }i = 1,...,k,
\end{equation}
where $\tau$ is the temperature hyperparameter. As the softmax temperature $\tau$ approaches 0, samples from the Gumbel-Softmax distribution become one-hot and the Gumbel-Softmax distribution becomes identical to the categorical distribution $p(z)$. 
% In our work we used an implementation trick that allows to work with one-hot vector even when the temperature is not near 0.

\section{Hyperparameters for experiments in \cref{tab:predictive-qual}}
\label{sec:hyperparams}

We perform a grid search using a held-out validation set to select the best performing hyperparameters for each model and dataset.  The hyperparameter settings used for each model and dataset are described below.

\textbf{Baseline MLE SDPP} \citep{gartrell2016bayesian}.  For this model, we use $K$ for the number of item feature dimensions for the symmetric component $\V$, and $\alpha$ for the regularization hyperparameter for $\V$.  We use the following hyperparameter settings:
\begin{itemize}
    \item All datasets: $K=30, \alpha=0$, $\text{batch-size}=200$.
\end{itemize}

\textbf{Baseline MLE NDPP} \citep{gartrell2019nonsym}.  For this model, to ensure consistency with the notation used in~\cite{gartrell2019nonsym}, we use $D$ to denote the number of item feature dimensions for the symmetric component $\V$, and $D'$ to denote the number of item feature dimensions for the nonsymmetric components, $\B$ and $\C$.  As described in~\cite{gartrell2019nonsym}, $\alpha$ is the regularization hyperparameter for the $\V$, while $\beta$ and $\gamma$ are the regularization hyperparameters for $\B$ and $\C$, respectively. We use the following hyperparameter settings:
\begin{itemize}
    \item All datasets: $D = D'=30, \alpha = \beta = \gamma = 0$, $\text{batch-size}=200$.
\end{itemize}

\textbf{WDPP (ours)}.  We use $K$ to denote the number of item feature dimensions for $\V$.  $\alpha$ is the regularization hyperparameter. $\tau_C$, $\tau_P$, $\tau_M$ and $\tau_B$ are the temperature hyperparameters for Cholesky-based DPP sampling, stochastic softmax trick (SST) Poisson sampling, SST multinomial sampling, and the SST Bernoulli sampling, respectively.  We use the following hyperparameter settings:
\begin{itemize}
    %\item Amazon apparel dataset: $\alpha=0.01$.
    %\item Amazon diaper dataset: $\alpha=0.01$.
    %\item Amazon feeding dataset: $\alpha=0.01$.
    \item All datasets: $K=30$, $\alpha=0.01$, $\tau_C =  \tau_P = 0.1$, $\tau_M = 1$, $\tau_B = 10^{-8}$, $\text{batch-size}=400$.
\end{itemize}

During WDPP training, we anneal both the learning rate and $\alpha$.

\section{Additional Experimental Results}
\label{sec:additional-experiments}

\begin{figure}[h]
    \centering
    {\includegraphics[width=6cm]{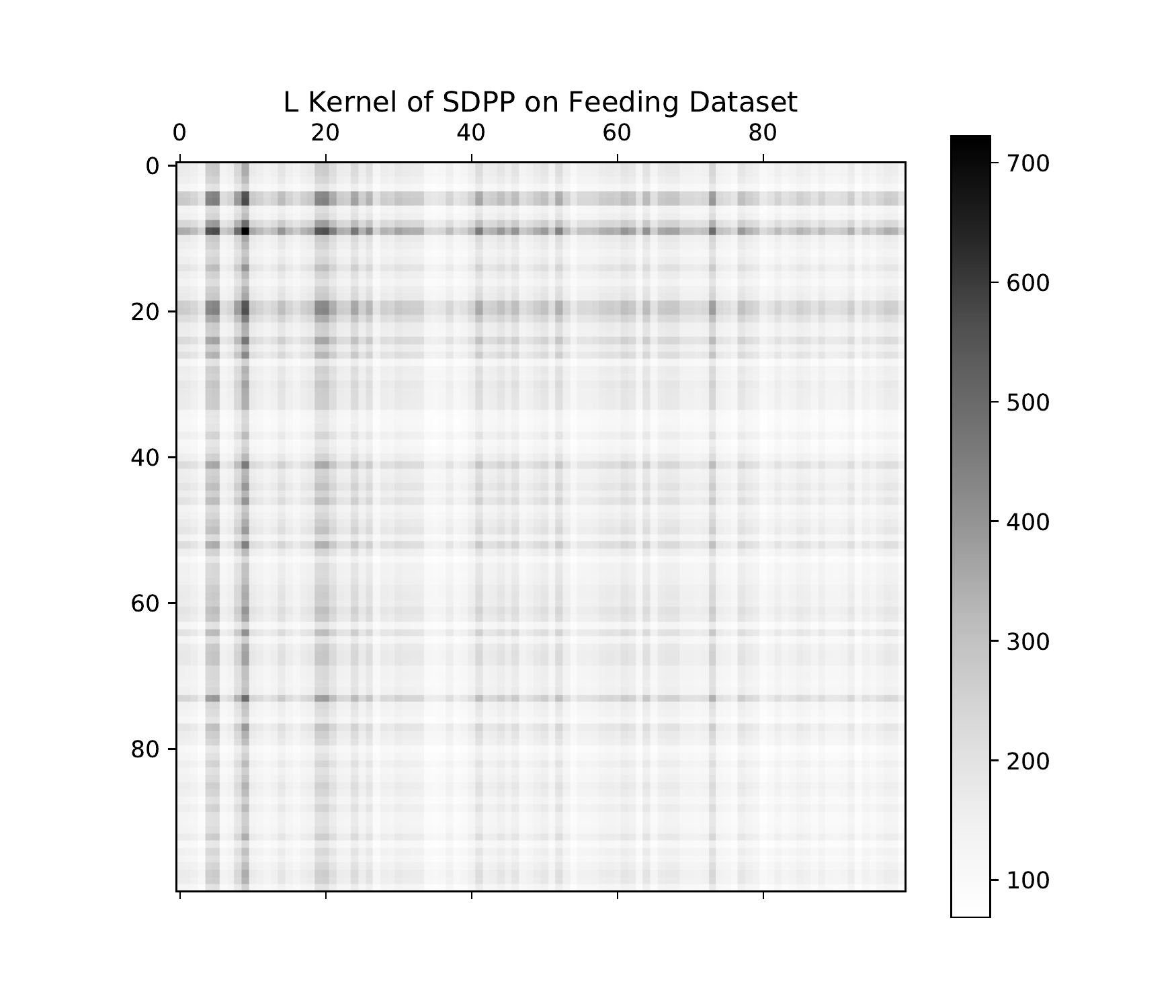}}
    {\includegraphics[width=6cm]{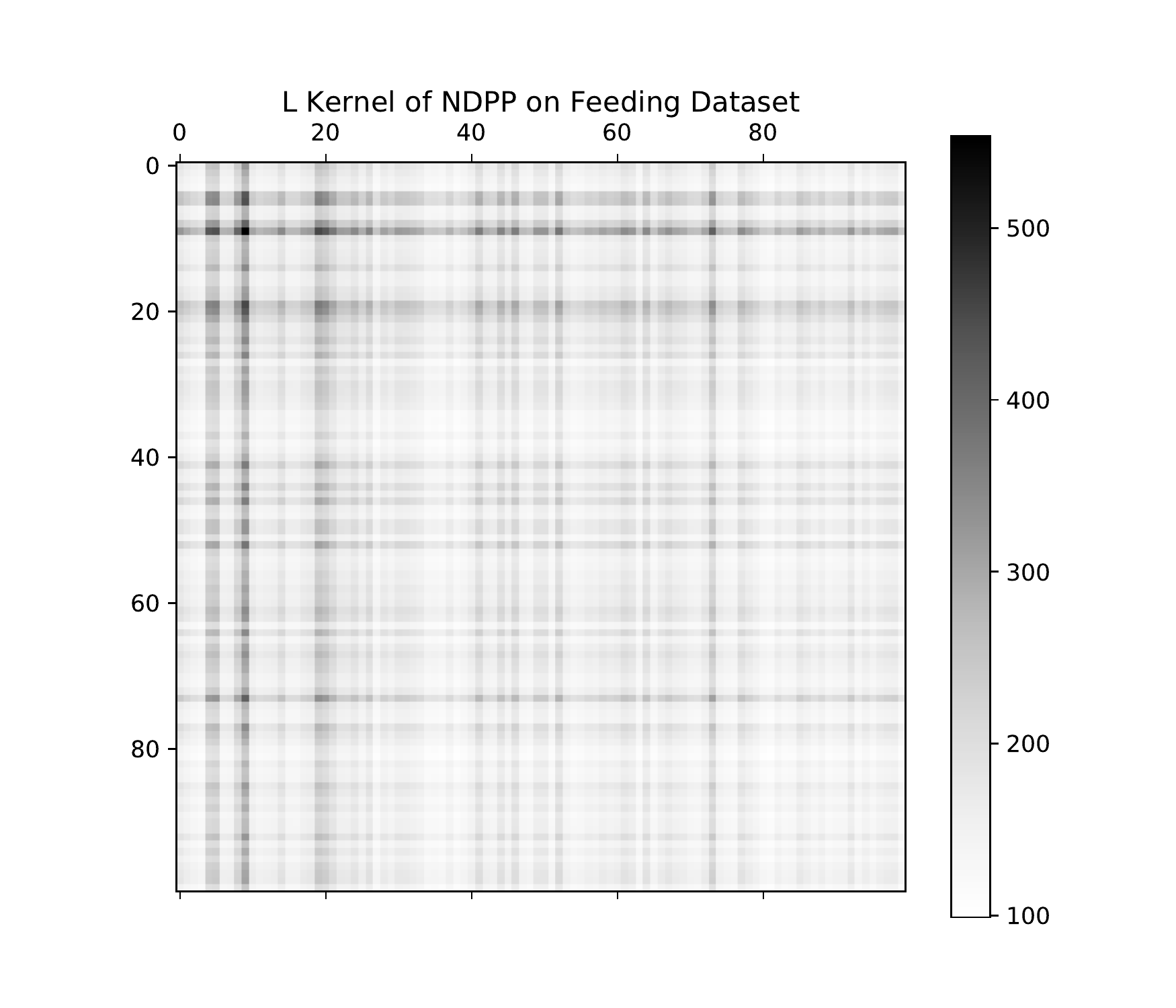}}
    {\includegraphics[width=6cm]{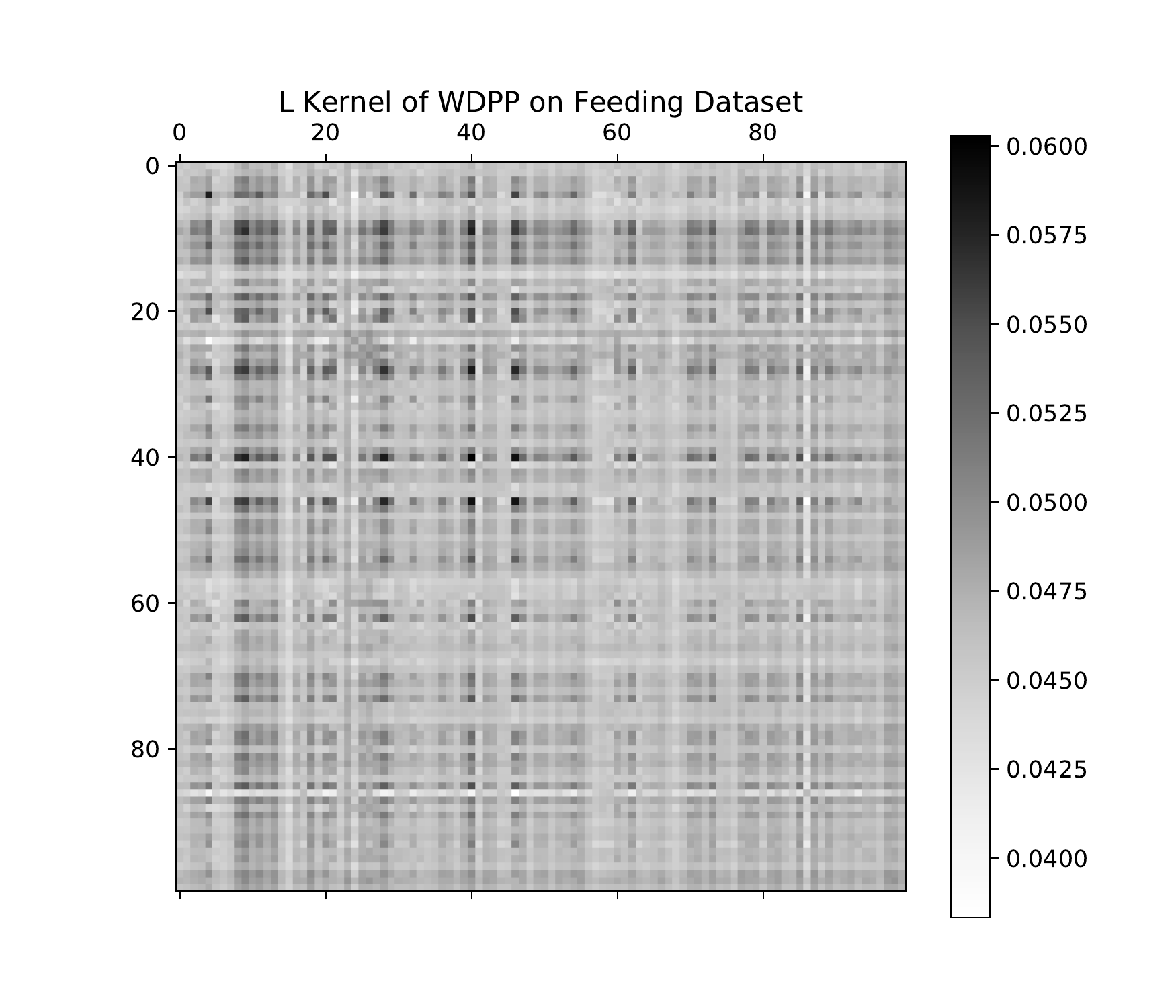}}
    \caption{Comparison of the learned DPP kernels for the MLE SDPP, MLE NDPP, and WDPP models, for the Amazon feeding dataset.}
    \label{fig:kernel-comparison}
\end{figure}

\begin{figure}[h]
    \centering
    {\includegraphics[scale=0.4]{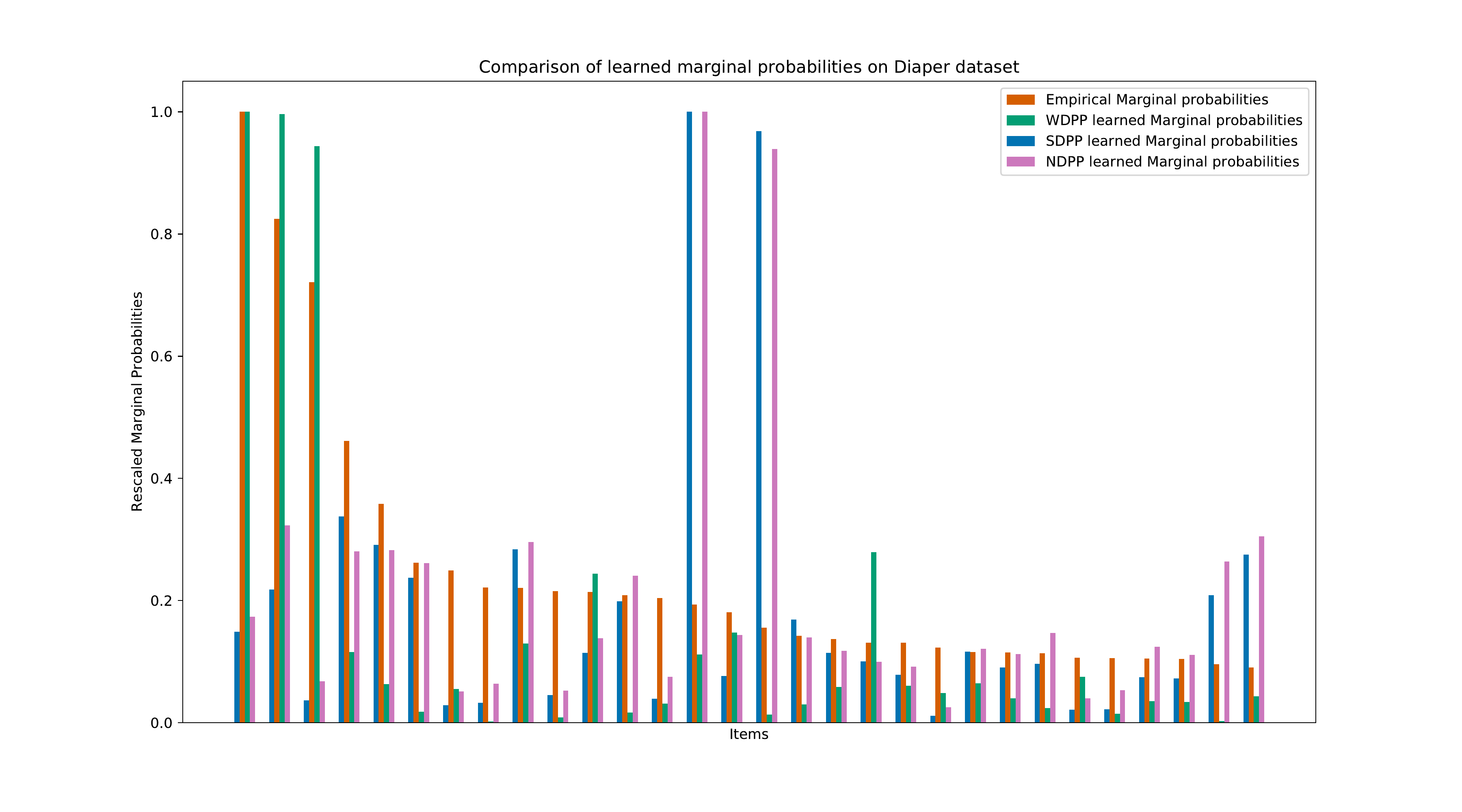}}
    \caption{Comparison of the empirical marginal probabilities to the learned marginal probabilities captured by the SDPP and WDPP models, for the Amazon diaper dataset.}
    \label{fig:marginals-comparison}
\end{figure}

\cref{fig:kernel-comparison} shows a plot of the kernels learned by the MLE SDPP and WDPP models for the Amazon feeding dataset.  We see more apparent structure in the WDPP kernel, suggesting that our Wasserstein learning approach allows the DPP to capture more structure from the data than when trained using MLE.  
In \cref{fig:marginals-comparison} we compare a portion of the empirical marginal item distribution with the marginals captured by the MLE SDPP and WDPP models when trained on the Amazon diaper dataset.  We see that WDPP appears to learn a better approximation of the true marginal distribution of the items in the data.  Finally, \cref{tab:most-represented-subsets} shows a collection of some of the most common non-singleton subsets (modes) from the test set, and samples generated by the WDPP, SDPP, and NDPP models, for the Amazon apparel dataset.  Compared to the DPPs trained by MLE, we see that our WDPP model generates subsets that are much closer to subsets found in the empirical test set.  

% To be sure that the structure is defining the true set distribution rather than noise, we can look at the marginal probabilities and compare the learning with the empirical distribution of the marginale probabilities of the sets (i.e. the popularity of the items).

% Now let's take a look at the most represented sets in the Test dataset of the Apparel dataset and compare them with the most represented sampled sets given by SDPP, NDPP and WDPP:

\begin{table}[h]
  \caption{Most common subsets in the empirical test set and samples generated by the WDPP, MLE SDPP, and MLE NDPP models, for the Amazon apparel dataset.}
    \begin{center}
    \scalebox{0.90}{
        \begin{tabular}{ |c|c|c|c| } 
        \hline
        \begin{tabular}{@{}c@{}}Most represented\\Test subsets\end{tabular}&\begin{tabular}{@{}c@{}}Most represented\\sampled subsets for WDPP\end{tabular}&\begin{tabular}{@{}c@{}}Most represented\\sampled subsets for SDPP \end{tabular}&\begin{tabular}{@{}c@{}}Most represented\\sampled subsets for NDPP\end{tabular}\\
        \hline
        (1, 12) & \bf (1, 12) & (1, 9) & (1, 9) \\
        (12, 23) & \bf (11, 12) & (9, 20) & (9, 20) \\
        (12, 22) & \bf (12, 26) & (9, 21) & (9, 64) \\ 
        (12, 50) & \bf (2, 12) & (9, 64) & (9, 28, 78) \\
        (2, 12) & \bf (12, 23) & (9, 88) & (9, 37) \\
        (12, 57) & \bf (12, 22) & (9, 28) & (9, 43) \\
        (12, 26) & \bf (12, 57) & (1, 8) & (9, 19) \\
        (11, 12) & (12, 39) & (9, 95) & (9, 49) \\
        (4, 12) & (3, 12) & (9, 66) & (9, 24) \\
        (31, 82) & (1, 22) & (9, 54) & (17, 28) \\
        \hline
        \end{tabular}
    }
    \end{center}
    \label{tab:most-represented-subsets}
\end{table}

\end{document}